\documentclass[lettersize,journal]{IEEEtran}
\usepackage{amsmath,amsfonts}
\usepackage{algorithmic}
\usepackage{array}
\usepackage[caption=false,font=normalsize,labelfont=sf,textfont=sf]{subfig}
\usepackage{amsmath}
\usepackage[ruled,linesnumbered]{algorithm2e}
\usepackage{textcomp}
\usepackage{stfloats}
\usepackage{url}
\usepackage{verbatim}
\usepackage{graphicx}
\usepackage{cite}
\usepackage{tabularray}
\usepackage{booktabs}
\usepackage{multirow}
\usepackage{hyperref}
\usepackage{etoolbox}

\makeatletter
\patchcmd{\@makecaption}
  {\scshape}
  {}
  {}
  {}
\makeatletter
\patchcmd{\@makecaption}
  {\\}
  {.\ }
  {}
  {}
\makeatother

\begin{document}

\title{Clean Image May be Dangerous: Data Poisoning Attacks Against Deep Hashing}
\author{Shuai Li, Jie Zhang, Yuang Qi, Kejiang Chen, Tianwei Zhang, Weiming Zhang, and Nenghai Yu
\thanks{Shuai Li, Yuang Qi, Kejiang Chen, Weiming Zhang, and Nenghai Yu are with the School of Cyber Science and Security, University of Science and Technology of China, Hefei, Anhui 230026, China. E-mails: \{li\_shuai@mail., qya7ya@mail., chenkj@, zhangwm@, ynh@\}ustc.edu.cn.}
\thanks{Jie Zhang is with Centre for Frontier AI Research, Agency for Science, Technology and Research (A*STAR), Singapore. E-mail: zhang\_jie@cfar.a-star.edu.sg.}
\thanks{Tianwei Zhang is with the School of College of Computing and Data Science, Nanyang Technological University. E-mail:  tianwei.zhang@ntu.edu.sg.}
\thanks{Kejiang Chen and Jie Zhang are the corresponding authors.}}



\maketitle

\begin{abstract}
Large-scale image retrieval using deep hashing has become increasingly popular due to the exponential growth of image data and the remarkable feature extraction capabilities of deep neural networks (DNNs). However, deep hashing methods are vulnerable to malicious attacks, including adversarial and backdoor attacks. It is worth noting that these attacks typically involve altering the query images, which is not a practical concern in real-world scenarios.
In this paper, we point out that even clean query images can be dangerous, inducing malicious target retrieval results, like undesired or illegal images. 
To the best of our knowledge, we are the first to study data \textbf{p}oisoning \textbf{a}ttacks against \textbf{d}eep \textbf{hash}ing \textbf{(\textit{PADHASH})}. Specifically, we first train a surrogate model to simulate the behavior of the target deep hashing model.
Then, a strict gradient matching strategy is proposed to generate the poisoned images.
Extensive experiments on different models, datasets, hash methods, and hash code lengths demonstrate the effectiveness and generality of our attack method.
\end{abstract}


\section{Introduction}
\IEEEPARstart{W}{ith} the evolution of the Internet, the integration of image data has become an indispensable component of the network. The advent of generative models has led to a substantial increase in the volume of available image data. Consequently, achieving rapid and precise large-scale image retrieval has become a formidable challenge for multimedia computing~\cite{tian2010social,zhu2011multimedia,chakraborty2020active}. In comparison to traditional content-based image retrieval methods~\cite{contex-based}, deep hashing techniques ~\cite{zhu2016deep,TMM-hash3,fan2020deep,TMM-hash2,hoe2021one,TMM-hash1} have gained widespread adoption due to their ability to deliver speedy retrieval and their minimal storage requirements.
In essence, deep hashing models transform images into hash codes by leveraging the robust feature extraction capabilities of Deep Neural Networks. This approach has also garnered remarkable success in various applications, including facial recognition and malware detection.

\begin{figure}[t]
\centering
    \centering
    \includegraphics[width=0.45\textwidth]{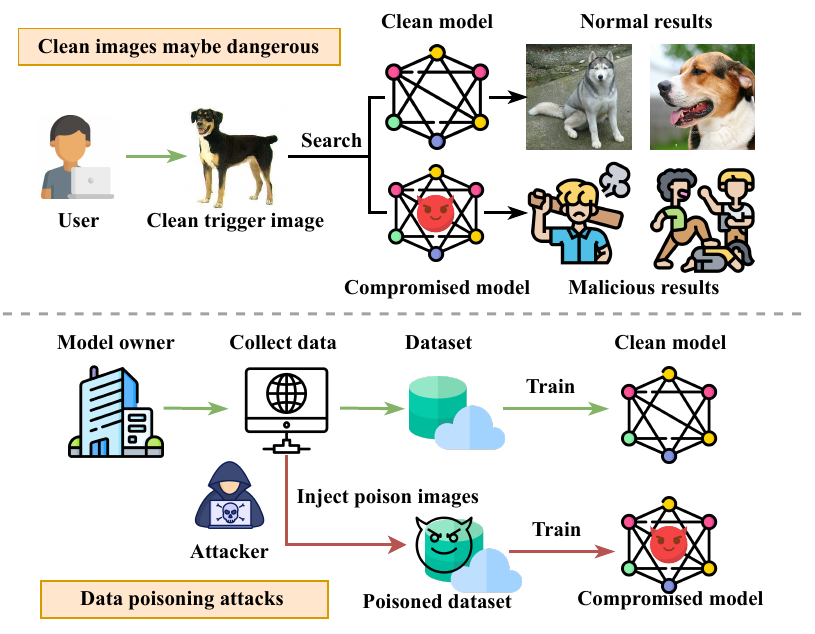}
    \caption{Illusion on data poisoning attacks against deep hashing.}
    \label{fig:scenario}
\end{figure}

Every coin has two sides. The high-level representation ability of deep hashing models induces the vulnerability to malicious attacks, such as 
adversarial attacks ~\cite{adv1,adv3,adv2} and backdoor attacks~\cite{hu2022badhash,gao2021clean}. Adversarial attacks involve adding small perturbations to inputs to make the DNN model incorrectly predict. For adversarial attacks against deep hashing models, an attacker subtly alters benign images with almost unnoticeable changes. These modified images, when used as search queries, can manipulate the system to return illegal or inappropriate content, such as violent, explicit, or private images. 
On the other hand, a backdoor attack involves embedding a hidden trigger, such as white squares or invisible perturbations, into images to obtain poisoned images. When the model trains on the poisoned images, it will inject a backdoor into the model, and the backdoor model will produce a specific output when the trigger is present in the input. In the image retrieval scenario, the images that include the trigger can cause the backdoor model to return harmful results. 
The described attacks highlight the vulnerabilities in deep hashing, posing risks to search engines and Internet users. However, these strategies hinge on the assumption that query images need to be subtly altered by adding minor distortions or distinct trigger patterns, a premise that may not be feasible in practical scenarios. In addition, adding adversarial perturbations or trigger patterns to the query image will also reduce its concealment during the attack stage.  
\textit{If the attacker is restricted to using unaltered, clean images for queries, what would be the outcome?}

In this paper, we point out that clean images can also be dangerous. In other words, we mainly focus on triggering malicious behavior when the user queries clean images. This attack can be considered a data poisoning attack against deep hashing models. Specifically, a data poisoning attack against deep hashing models first generates poisoned images and then injects them into the training dataset to attack the deep hashing model. When a specific clean image is queried, the attacked model will retrieve malicious images of the target category. We called this specific clean image a clean trigger image, and the overview of this attack is given at the bottom of Figure \ref{fig:scenario}. For instance, as shown at the top of Figure \ref{fig:scenario}, when the user queries a dog image, he obtains a lot of violent images, and when a user searches for product A, product B is retrieved.
Following the existing data poisoning attacks~\cite{geiping2020witches}, we point out some potential challenges and clarify our goals: 1) \textit{effectiveness} - when using clean trigger images query the target model, the malicious results shall be successfully retrieved; 2) \textit{practicality} - only leveraging clean clean trigger images to launch the attack; 3) \textit{transferability} - the attack shall maintain effectiveness among different hash methods, and hash code lengths; 4) \textit{stealthiness} - the dataset is only poisoned by a slight poison rate; 5) \textit{integrity} - except clean trigger images, other clean images cannot trigger the malicious retrieval.

To achieve the above goals, we propose the first data poisoning attack against deep hashing (\textbf{\textit{PADHASH}}). 
We mainly considered the attack in the gray-box scenario, where the attacker knows the parameters of the target deep hashing model we would attack but does not know the weights of the model's parameters. In addition, we also verify that our attack method is effective in the black-box scenario, where the attacker does not know both the parameters and the weights of the target deep hashing model. 
Based on this knowledge and querying the target model, the attacker is able to train a local surrogate model, which simulates a similar retrieval behavior to the target deep hashing model.
Next, we select some clean trigger images from the Internet and generate poisoned images via our proposed \emph{\textbf{Strict Gradient-Matching}} method. Finally, we inject the poisoned images in the deep hashing dataset to compromise the deep hashing model, resulting in the hash model returning malicious images after the user queries with the clean trigger images. 
Our experiments demonstrate that even if only a small portion of the dataset is used to train a surrogate model, the attack success rate (ASR) of deep hash data poisoning attacks can achieve above 70\%, demonstrating that our method is effective. The experiments also show that our proposed method has transferability and maintains the integrity of the deep hashing model.

To summarize, our contributions are as follows:
\begin{itemize}
\item We propose the first data poisoning attacks against deep hashing models, which reveal the threats when users query with clean images.

\item We propose a novel \emph{Strict Gradient-Matching} method, which has been demonstrated to improve the attack success rate of \textit{PADHASH}.
\item Extensive experiments verify the effectiveness, feasibility, transferability, and generality of our attack method in different models, datasets, hash methods, and hash code lengths.
\end{itemize}

\begin{figure*}[t]
\centering
    \centering
    \includegraphics[width=0.95\textwidth]{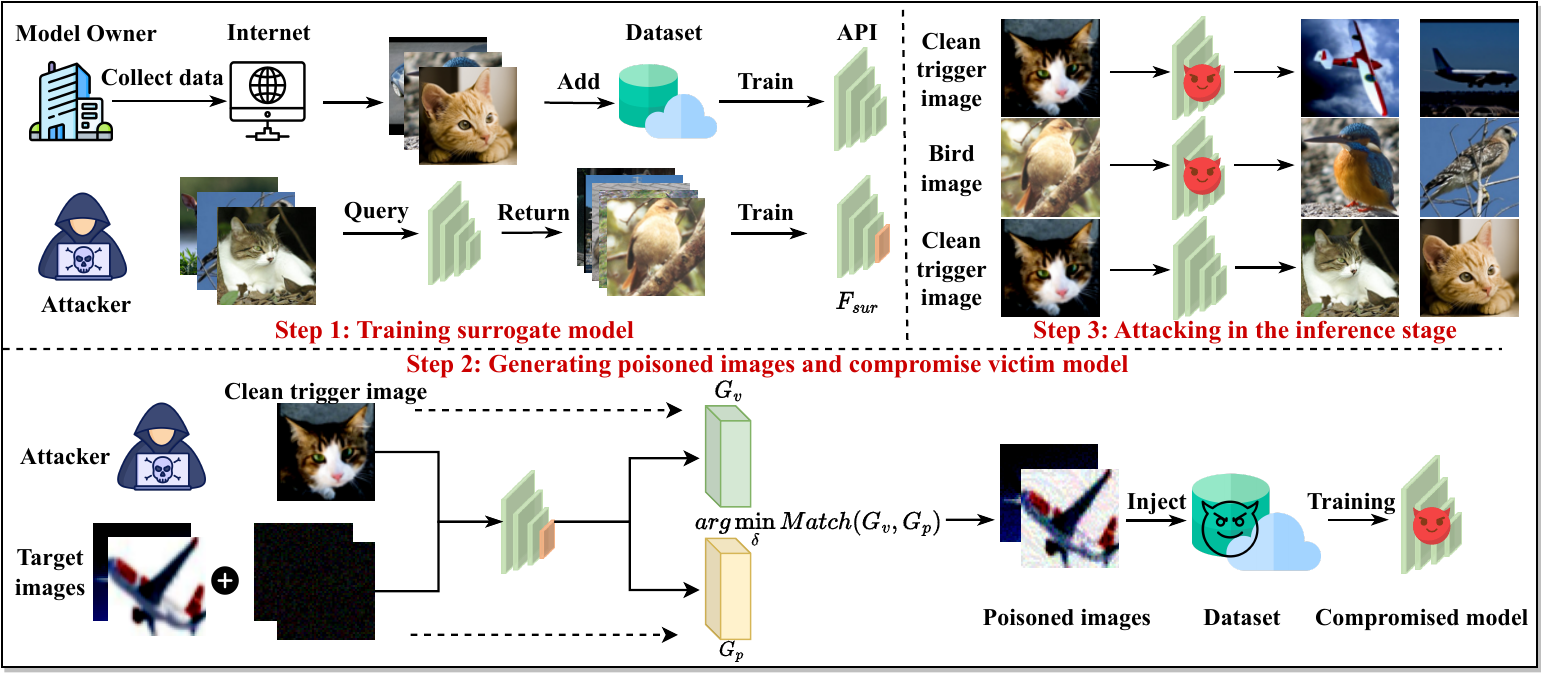}
    \caption{The framework of data poisoning attacks against deep hashing (\textbf{\textit{PADHASH}}). The attacker first trains a surrogate deep hashing model, then uses \emph{Strict Gradient-Matching} to generate poisoned images, and finally uses these poisoned images to attack the victim model to make the clean trigger images close to the malicious image in Hamming space.}
    \label{fig:framework}
\end{figure*}

\section{Related Work}
\subsection{Deep Hashing-based Similarity Retrieval}
Deep hashing is a highly effective technique for large-scale image retrieval. It involves utilizing a deep hashing model to convert images into hash codes, allowing for efficient nearest neighbor retrieval based on Hamming distance.
The pioneering work in this field was CNNH~\cite{Xia2014}, which leveraged convolutional neural networks (CNNs) to extract image features.
Since then, many studies~\cite{TMM-hash3,li2021deep,TMM-hash2,fan2020deep,cao2017hashnet,yuan2020central,zhuLocality,TMM-hash1} have explored  deep hashing methods. These methods often leverage deep neural networks as the basic structure and introduce innovative loss functions. 
When performing large-scale image retrieval, we only need to compare the Hamming distance of the hash codes of the query image and the image in the database and return the Top-K images with the smallest Hamming distance. Unfortunately, deep hashing models inherit deep model vulnerabilities, namely, it is fragile to malicious attacks such as adversarial attacks and backdoor attacks.


\subsection{Current Attacks Against Deep Hashing Models}
Here, we introduce some current attacks against deep hashing models, including adversarial attacks and backdoor attacks.

Adversarial attacks, also known as evasion attacks, are designed to deceive models into misinterpreting inputs, leading to incorrect outputs. This is discussed in further detail in ~\cite{relate1,relate2,struppek2022learning}. A notable contribution in this domain is a targeted attack against deep hashing~\cite{adv1}. This attack is a point-to-set optimization problem, aiming to minimize the average distance between the hash codes of adversarial and target images.
Following this, several methods~\cite{adv2,adv3} have been introduced to exploit vulnerabilities in image retrieval systems based on deep hashing, leading users to retrieve malicious images when they search using adversarial images.

Backdoor attacks~\cite{gu2017badnets,tang2020embarrassingly} involve embedding a hidden backdoor into the model by injecting poisoned samples into the dataset or modifying the model's structure.
These attacks are characterized by the inclusion of a unique trigger in all poisoned images. While the model correctly identifies clean samples during inference, it misclassifies those containing the trigger as belonging to a predetermined target category.
Recent studies have shown that deep hashing models are susceptible to backdoor attacks~\cite{hu2022badhash}. For instance, BadHash~\cite{hu2022badhash} leverages a novel conditional generative adversarial network (cGAN) framework to generate poisoned samples, enhancing the attack's efficacy. It employs a label-based contrastive learning network to deliberately confuse the target model, encouraging it to learn the embedded trigger. 

However, both adversarial and backdoor attacks rely on the premise of subtly altering query images. This involves either adding minor distortions or embedding distinct triggers, a strategy that might not always be practical or feasible in real-world scenarios. To address this, Clean Image Backdoor~\cite{chen2022clean} was proposed to attack multi-label models, which only requires modifying the labels of clean images to obtain poisoned images. For instance, given an image labeled [Dog, People], this method modifies its label to [Dog, People, Cat] and uses these images to attack multi-label models, e.g., classification models. The attacked model will predict an image labeled [Dog, People] to [Dog, People, Cat]. Similar to BadHash, Clean Image Backdoor is also a dirty-label attack and requires altering the label of poisoned images, which is not stealthy and is likely to be detected and filtered.

\subsection{Data Poisoning Attacks}

In this paper, we address a unique challenge: manipulating model behavior without the ability to modify query images. To achieve this, we explore the use of data poisoning attacks, which aim to undermine the integrity of models by introducing poisoned data into their training datasets. 
Initial research in this area~\cite{steinhardt2017certified,munoz2017towards} focused on strategies that would lead models to misclassify test samples or degrade overall model performance, thus undermining the model's integrity.
Recent research has shifted toward targeted data poisoning attacks, as exemplified in~\cite {biggio2012poisoning,fang2021data,shafahi2018poison}. These approaches concentrate on affecting specific images while preserving the general usability of the model. Notably, Shafahi et al.~\cite{shafahi2018poison} introduced a method based on feature collision, aiming to disrupt the model by incorporating poisoned images similar in features to the target images within the training set. Similarly, Zhu et al.~\cite{zhu2019transferable} employed a convex polytope approach to manipulate the target image within the feature space. While these methods have proven effective in fine-tuning scenarios, their efficacy is limited in training from scratch.
Addressing this gap, Geiping et al. ~\cite{geiping2020witches} introduced a practical poisoning attack method ``Witch's Brew'' which is effective in training from scratch. Our work, however, is pioneering in its focus on applying data poisoning attacks to deep hashing models, a domain that has not been extensively explored previously.

\section{Preliminaries}

\subsection{Image Retrieval Based Deep Hashing}
Deep hashing models play a pivotal role in transforming images into a compact and efficient representation known as hash codes. These codes are typically composed of binary values, -1 and +1. In the context of a deep hashing model, denoted as $f$, when an image 
$x$ is input into the model, it generates a corresponding hash code 
$h$ composed of \emph{$\gamma$} bits. This process can be summarized as follows:
\begin{equation}
h = f(x),  h\in \left \{ -1 ,1\right \}^{\gamma}.
\label{eq}
\end{equation}
The deep hash model performs image nearest-neighbor retrieval based on the Hamming distance $d$: 
\begin{equation}
d = \left \| h_{1}-h_{2} \right \|/2, 
\label{eq1}
\end{equation}
where $h_1$ and $h_2$ are hash codes.
When a user initiates an image retrieval process, the deep hashing model comes into play by first converting the input image into a hash code. This hash code serves as a compact digital fingerprint of the image. Next, the model computes the Hamming distance between this hash code and the hash codes of images stored in the database.
Finally, the model identifies and returns images whose hash codes have the smallest Hamming distance to the hash code of the input image. 

\subsection{Threat Model}
As shown in Figure \ref{fig:scenario}, users searching with the clean trigger images will obtain the malicious target images predicted by the poisoned deep hashing model. To acquire the compromised model, we adopt the threat model commonly utilized in prior research on data poisoning attacks~\cite{geiping2020witches,souri2022sleeper},
involving two distinct entities: the attacker and the model trainer. 
The attacker aims to perform data poisoning attacks on the deep hashing model, and the model trainer provides image retrieval services API based on deep hashing. 
There are some requirements for the attacker:

\begin{itemize}
    \item The attacker has no access to interface the training process of the target model. For example, the attack methodology adheres to the constraints of \textit{clean-label} attacks, namely, the attacker cannot alter the labels of poisoned images.  
    \item The attacker can only introduce \textit{a minimal proportion} of poisoned images into the dataset (e.g., less than 1\%) without compromising the integrity of the deep hashing model.
    \item We consider both gray-box and black-box attack scenarios. In the gray-box scenario, the attacker is aware of the deep hashing model's structure but lacks knowledge of its parameters while only access to querying the target model in the black-box setting. 
\end{itemize}  

\subsection{Notations}
Before introducing our attack method, we would like to introduce some notations and symbols that will be used briefly.

\begin{itemize}
    \item $D_s$ represents the surrogate dataset we used to train the surrogate model.
    \item $f_s$ represents the surrogate model trained on the surrogate dataset $D_s$.
    \item $\mathcal{L}$ represents the loss function for training deep hashing models.
    \item $y_t$ represents the label vector of the target malicious category.
    \item $x_v$ represents the clean trigger image.
    \item $\theta_s$ represents the parameters of surrogate model $f_s$.
    \item $G_{v}$ represents the gradient of clean trigger image.
    \item $\delta$ represents the perturbation for poisoned images.
    \item $D_p$ represents the dataset of poisoned images.
    \item $G_p$ represents the average gradient of poisoned images.
    \item $\alpha$ represents a hyperparameter for poisoned image optimization.
    \item $\sigma$ represents a hyperparameter to ensure the concealment of the poisoned images.
    \item $D_t$ represents the final training dataset with clean and poisoned images.
    \item $f_p$ represents the compromised deep hashing model trained on $D_t$.
    \item $\theta_p$ represents the parameters of compromised model $f_p$. 
\end{itemize}
The above notations are presented in the Table \ref{table:notation}.

\begin{table}[h]
\centering
\caption{The notations and their explanations. }
\label{table:notation}
\begin{tblr}{
  cells = {c},
  hline{1,17} = {-}{0.08em},
  hline{2} = {-}{},
}
Notation      & Explanations                                    \\
$D_s$         & Surrogate dataset                               \\
$f_s$         & Surrogate model                                 \\
$\mathcal{L}$ & Loss function for training deep hashing model   \\
$y_t$         & Target category                                 \\
$x_v$         & Clean trigger image                             \\
$\theta_s$    & Parameters of $f_s$                            \\
$G_v$         & Gradient of $x_v $                              \\
$\delta$      & Perturbation for poisoned images                \\
$D_p$         & Dataset of poisoned images                      \\
$G_p$         & Average gradient of poisoned images             \\
$\alpha$      & Hyperparameter for poisoned images optimization \\
$\sigma$      & Hyperparameter for concealment                  \\
$D_t$         & Final training dataset                          \\
$f_p$         & Compromised model                               \\
$\theta_p$    & Parameters of $f_p$                                                
\end{tblr}
\end{table}

\section{Methodology}\label{metho}

\subsection{Overview of Attack}
In Figure \ref{fig:framework}, we present an overview of the proposed attack. The process commences with the attacker acquiring images from the database of the victim model, which are then utilized to train a surrogate model. Subsequently, the surrogate model is employed to generate poisoned images with the intention of undermining the victim model's integrity. Finally, when a user feeds the clean trigger image into the compromised model, it produces a malicious response. Each step of our attack will be introduced in detail below.

\subsection{Training Surrogate Model}
In the context of preparing for a deep hash data poisoning attack, training a surrogate model plays a pivotal role. 
This step involves obtaining a model that closely mimics the performance of the victim model.
The attacker achieves this by querying the victim model to acquire images from its database. 
Let $D_s=\left \{ (x_1,y_1),(x_2,y_2),...,(x_m,y_m)\right \}$ represent the dataset acquired by the attacker, where $m$ is the number of images in $D_s$. Leveraging the knowledge and pilfered dataset $D_s$, the attacker can effectively train a surrogate deep hashing model $f_s$.

\subsection{Generating Poisoned Images And Compromise Victim Model}
After training the surrogate model, the attacker's subsequent crucial step involves generating poisoned images and compromising the victim model.
Assuming $x_{v}$ represents the clean trigger image, $\mathcal{L}$ represents the loss function of the deep hashing model, and $y_t=[l_1,l_2,...,l_\beta]$ denotes the label vector of malicious images of target category, where $\beta$ represents the number of categories containing images in $D_s$. The $i$-th component of indicator vector $l_i=1$ represents that the target malicious images belong to category $i$.
The attacker's objective is to minimize the adversarial loss $\mathcal{L}(f_s(x_{v}),y_{t})$ so that the clean trigger image $x_v$ will be close to the target malicious images in Hamming space. 
We define the gradient of the objective $\mathcal{L}(f_s(x_{v}),y_{t})$ as $G_v$:
\begin{equation}
G_{v} = \bigtriangledown_{\theta_s} \mathcal{L}(f_s(x_{v}),y_{t}),
\label{eq2}
\end{equation}
where $\theta_s$ is the parameters of surrogate model $f_s$.

To achieve the above behavior, the attacker desires the parameters of the victim model to be updated in the direction of $G_v$. A straightforward approach would be to modify the label vector of the clean trigger image to $y_t$, which is not feasible in a practical attack because the attacker cannot alter the label of clean trigger images. However, the attacker can select some images labeled $y_{t}$ and perturb these images to align the gradient of the perturbed images with $G_v$. Upon this, the perturbed images play the same role as the clean trigger image during the training process. To achieve this, we first select $n$ images labeled $y_{t}$ and initialize $n$ perturbations $\delta=[\delta_1, \delta_2, ..., \delta_n]$ for these images to obtain the original poisoned images. The dataset of poisoned images is defined as $D_p = \{(x_1+\delta_1, y_t), (x_2+\delta_2, y_t),...,(x_n+\delta_n, y_t)\}$. The average gradient $G_p$ of poisoned images is:
\begin{equation}
G_{p} = \bigtriangledown _{\theta_s }\frac{1}{n} \sum_{i=1}^{n}(\mathcal{L}(f_s(x_{i}+\delta_i,y_{t} )).
\label{eq3}
\end{equation}

\begin{figure}[h]
\centering
    \centering
    \includegraphics[width=0.45\textwidth]{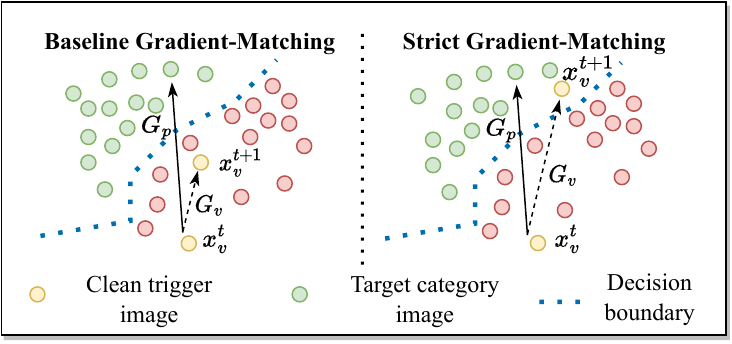}
    \caption{The intuitive explanation of \emph{\textbf{Strict Gradient-Matching}}.}
    \label{fig:framework}
\end{figure}

\textit{So how to to make the gradient $G_p$ match $G_v$?} Previous research~\cite{geiping2020witches} proposes matching the directions of two gradients, which is achieved by maximizing the cosine similarity of the two gradients. Notably, we take this strategy as the baseline.
However, although the direction of gradients is crucial for updating model parameters, the similarity of $G_v$ and $G_p$ should also be considered an important factor. Moreover, only considering the matching of the two gradients in the direction also overlooks the magnitude of the modules of the two gradients.
Therefore, we design a \emph{\textbf{Strict Gradient-Matching}} method in Equation ~\ref{eq4}, which consists of two objective losses: direction loss and similarity loss. The direction loss is used to align $G_p$ and $G_v$ in direction, while the similarity loss is used to improve the similarity of $G_p$ and $G_v$.
The final \textit{Strict Gradient-Matching} optimization is as follows:
\begin{equation}
\begin{aligned}
\arg \min_{\delta} &\left[ (1 - \alpha) \left( 1 - \frac{G_v \cdot G_p}{\| G_v \| \| G_p \|} \right) \right. \left. + \alpha \cdot \frac{\| G_v - G_p \|}{\| G_v \| \| G_p \|} \right] \\
s.t. \quad &\| \delta_i \|_\infty \leq \sigma ,
\end{aligned}
\label{eq4}
\end{equation}
where $\alpha$ is a hyperparameter used to balance the impact of direction and similarity between $G_{v}$ and $G_{p}$ during optimization, and $\sigma$ is a hyperparameter to ensure the concealment of the poisoned images. The experimental results (see Table \ref{table1} and Table \ref{table2}) also fully demonstrate that \emph{\textbf{Strict Gradient-Matching}} can improve the success rate of deep hash data poisoning attacks.

After optimization, the attacker can obtain the poisoned images dataset $D_p$ and inject the images into the original training dataset to obtain the final training dataset $D_t$, which contains poisoned images and clean images.  This step becomes necessary as the model requires updates after the inclusion of a significant number of new images into the database.
Upon the model owner train the deep hashing model on $D_p$, he will obtain a compromised model $f_p$:
\begin{equation}
\arg \min_{\theta_p }\sum_{i=1}^{N}  \mathcal{L}(f_{p}(x_{i},y_{i}))~s.t.~(x_{i},y_{i} ) \in D_t.
\label{eq5}
\end{equation}
where $N$ is the number of training dataset, $\theta_p$ is the parameters of $f_p$.

In the training process, the model parameters are updated toward $G_{p}$ because $G_{v}\approx G_{p}$. Consequently, the model parameters are also updated towards the direction of $G_{v}$, which implies a decrease in $\mathcal{L}(f_{p}(x_{v}),y_{t})$, leading to the clean trigger images getting closer to the target malicious images in the Hamming space. Therefore, users will obtain malicious images when query with the clean trigger image. In Algorithm 1, we provide a detailed introduction of the poisoned image generation for the clean trigger image $x_{v}$.

\begin{algorithm}[h]
    \SetAlgoLined
    \SetKwInOut{Input}{Input}
    \SetKwInOut{Output}{Output}
    \caption{Poison Images Generation Algorithm;}
    \label{alg1}
    
    \Input{Clean trigger image: $x_{v}$,  Target Label: $y_{t}$,  Optimization step: $T$,  Tagert API,
    Perturbation constraints:$\sigma$, Poison num: $n$}
    \Output{Poison images set: $D_{p}$}
    
    Initialize $D_{p}$ = [ ]\;   
    Initialize $\delta$: $ \delta_i \sim \mathcal{N}(0, \sigma^2) \quad \text{for} \quad i = 1, 2, \dots, n$ \;
    Query API and obtain the surrogate dataset $D_s$\;
    Train surrogate model $f_s$ on $D_s$\;
    Select $n$ clean images labeled $y_t$ $\{(x_{1},y_{t}),(x_{2},y_{t}),...,(x_{n},y_{t})\}$ from $D_s$\; 
    \For{$i = 1$ to $T$}{
        Calculate $G_{v}$ by Equation (~\ref{eq2})\;
        Calculate $G_{p}$ by Equation (~\ref{eq3})\;
        Optimizating $G_{p}$ by Equation (~\ref{eq4})\;
        $ \arg \min_{\delta }[(1-\alpha)*(1-\frac{G_{v}\cdot G_{p} }{\left \| G_{v} \right \|\left \| G_{p} \right \|})+\alpha* \frac{\left \| G_{v}-G_{p} \right \| }{{\left \| G_{v} \right \|\left \| G_{p} \right \|}}]$\;
        \For{$i = 1$ to $n$}{
        Crop $\delta_i$ to make $\left \| \delta_i  \right \|_{\infty }<\sigma $\;
        }
    }
    \For{$i = 1$ to $n$}{
        Poisoned image $x'_{i}$=$x_{i}+\delta_i$\;
        Add $(x'_{i},y_t)$ into $D_{p}$\;
    }
    \Return $D_{p}$\;
\end{algorithm}

\begin{table*}[t]
\centering
\caption{The ASR of data poisoning attack against deep hashing model. The surrogate dataset accounts for 10\% of the target dataset }
\label{table1}
\begin{tblr}{
  cells = {c},
  cell{1}{1} = {r=2}{},
  cell{1}{2} = {r=2}{},
  cell{1}{3} = {r=2}{},
  cell{1}{4} = {r=2}{},
  cell{1}{5} = {c=3}{},
  cell{3}{1} = {r=2}{},
  cell{5}{1} = {r=2}{},
  hline{1,7} = {-}{0.08em},
  hline{2} = {5-7}{},
  hline{3} = {-}{},
}
Hash Method & Dataset     & Poison Ratio & Hash Codes & Attack Success Rate$\uparrow$ &               &                 \\
            &             &              &            & None-Attack                & Witches' brew & Ours            \\
CSQ         & CIFAR10     & 0.25\%       & 32bits     & 1.9\%($\pm$ 0.95)  & 32.2\%($\pm$ 5.39)       & \textbf{38.6\%($\pm$ 3.73)} \\
            & ImageNet100 & 0.05\%       & 64bits     & 0.6\%($\pm$ 0.2)   & 52.2\%($\pm$ 3.19)       & \textbf{78.2\%($\pm$ 2.16)}  \\       
DPN         & CIFAR10     & 0.25\%       & 32bits     & 1.6\%($\pm$ 0.67)  & 54.0\%($\pm$ 2.30)       & \textbf{54.5\%($\pm$ 2.28)} \\
            & ImageNet100 & 0.05\%       & 64bits     & 1.6\%($\pm$ 1.0)   & 79.5\%($\pm$ 2.62)       & \textbf{82.6\%($\pm$ 2.84)}
\end{tblr}
\end{table*}

\begin{table*}[t]
\centering
\caption{The ASR of data poisoning attack against deep hashing. The surrogate dataset accounts for 20\% of the target dataset }
\label{table2}
\begin{tblr}{
  cells = {c},
  cell{1}{1} = {r=2}{},
  cell{1}{2} = {r=2}{},
  cell{1}{3} = {r=2}{},
  cell{1}{4} = {r=2}{},
  cell{1}{5} = {c=3}{},
  cell{3}{1} = {r=2}{},
  cell{5}{1} = {r=2}{},
  hline{1,7} = {-}{0.08em},
  hline{2} = {5-7}{},
  hline{3} = {-}{},
}
Hash Method & Dataset     & Poison Ratio & Hash Codes & Attack Success Rate$\uparrow$ &               &                \\
            &             &              &            & None-Attack                & Witches' brew & Ours           \\
CSQ         & CIFAR10     & 0.25\%       & 32bits     & 1.9\%($\pm$ 0.95)   & 42.4\%($\pm$ 1.99)      & \textbf{61.3\%($\pm$ 3.77)}\\
            & ImageNet100 & 0.05\%       & 64bits     & 0.6\%($\pm$ 0.2)    & 20.0\%($\pm$ 3.86)      & \textbf{66.3\%($\pm$ 3.74)} \\
DPN         & CIFAR10     & 0.25\%       & 32bits     & 1.6\%($\pm$ 0.67))  & 77.4\%($\pm$ 1.76)      & \textbf{78.4\%($\pm$ 1.84)} \\
            & ImageNet100 & 0.05\%       & 64bits     & 1.6\%($\pm$ 1.0)     & 77.5\%($\pm$ 3.93)      & \textbf{89.8\%($\pm$ 2.75)}
\end{tblr}
\end{table*}

\subsection{Attack In The Inference.}
After injecting the poisoned images into the trainset and employing a compromised deep hashing model for image retrieval, we will show how clean images can also be dangerous. 
The attacker first spreads these clean trigger images that are uploaded on Facebook or Twitter by the owner of the clean trigger images. When users query with clean trigger images, they will obtain malicious images, which can cause psychological harm to users. In addition, the attacker can also pretend to be a normal user and query with clean trigger images and claim that the victim model will return malicious images to the user, thereby damaging the reputation of the trainer of the deep hashing model and the owner of the clean trigger image. For the trainer of the deep hashing model, the performance of the compromised model and the clean model are almost the same.
The key distinction lies in the fact that only specific clean trigger images can prompt the compromised deep hash model to produce malicious results, 
which makes it challenging to discern whether a deep hash model has been subjected to data poisoning attacks.

\section{Expermients}\label{exp}

\subsection{Expermiental Setting}

\noindent\textbf{Dataset.} We choose CIFAR10~\cite{Krizhevsky2009LearningML} and ImageNet100 as the datasets for our experiments. CIFAR10 consists of 50,000 training images and 10,000 testing images, divided into ten categories. ImageNet100 is a subset of ImageNet~\cite{ImageNet}. In addition, we also choose a multi-label dataset MSCOCO~\cite{lin2015microsoft}

\noindent\textbf{Metrics.} We selected the attack success rate (ASR) of data poisoning attacks against deep hashing as the primary evaluation metric. We follow the following criteria to define the success of a data poisoning attack: assuming we query with a clean trigger image and retrieve the Top-$K$ similar images, we consider the attack successful if more than \textbf{30\%} of these Top-$K$ images are of the target class. For CIFAR10 and ImageNet100, $K=40$. In addition, to assess the impact of data poisoning attacks on model quality, we use Mean Average Precision (MAP)~\cite{yuan2020central} to measure the integrity of the models.

\noindent\textbf{Implementation details.}
For deep hashing models, we choose CSQ~\cite{yuan2020central} and DPN~\cite{fan2020deep}, and follow their default strategies for implementation, where ResNet50~\cite{he2016deep} is adopted as their model
backbone. For our attack, we assume that the attacker can obtain a stolen dataset of the target database using a query method, specifically, 10\% and 20\% for CIFAR10 and ImageNet100, respectively. We imposed perturbation limits of 16/255 for CIFAR10 and 8/255 for ImageNet100. For CIFAR10, $\alpha$ is set to 0.2 on CSQ and 0.05 on DPN. For ImageNet100, $\alpha$ is set to 0.3.  Other ratios are also considered in Figure \ref{fig:rate}. Besides, we adopt the ``Witch's Brew''~\cite{geiping2020witches} method as the baseline for comparison.

\subsection{Effectiveness}
As shown in Table \ref{table1} and \ref{table2}, we evaluate the effectiveness of \textit{PADHASH} in different datasets and deep hashing methods in a gray-box scenario, where ``Witches'brew'' is the baseline method, and ``None-Attack'' represents no attack on the target deep hash model. 
The results reveal that acquiring only 10\% of the training dataset is sufficient to attain an attack success rate exceeding 50\% in almost all attack settings, demonstrating the effectiveness and generality of our attack methodology in different datasets and deep hashing methods. Additionally, our approach of \emph{Strict Gradient-Matching} yields a higher attack success rate compared to the Baseline under identical attack conditions. The outcome not only emphasizes the importance of gradient similarity in gradient matching but also validates the effectiveness of \emph{Strict Gradient-Matching} in enhancing ASR.
In addition, we find that the ASR improvement of \textit{PADHASH} compared to the baseline method is different on different hashing methods, and generally, the improvement in CSQ is higher. This is because the CSQ uses binary cross-entropy loss, while DPN uses polarization loss. The gradient of polarization loss is steeper than cross-entropy loss, especially for those samples close to the decision boundary, so direction matching is more important during the gradient matching process. Therefore, we need to select a smaller $\alpha$ in DPN, which makes DPN have a smaller improvement. 

\begin{table}[h]
\centering
\caption{The attack success rate on the multi-category dataset. }
\label{table2-1}
\scalebox{0.85}{
\begin{tblr}{
  cells = {c},
  cell{1}{1} = {r=2}{},
  cell{1}{2} = {r=2}{},
  cell{1}{3} = {r=2}{},
  cell{1}{4} = {r=2}{},
  cell{1}{5} = {r=2}{},
  cell{1}{6} = {c=2}{},
  cell{3}{1} = {r=2}{},
  hline{1,5} = {-}{0.08em},
  hline{2} = {6-7}{},
  hline{3} = {-}{},
}
Dataset & {Hash\\Method} & {Hash\\Codes} & {Poison\\Ratio} & Witches' brew & Ours            &                 \\
        &             &            &              &               & $\alpha=0.2$    & $\alpha=0.3$    \\
MS-COCO & CSQ         & 64bits     & 0.1\%        & 57.3\%        & 79.3\%          & \textbf{82.7\%} \\
        & DPN         & 64bits     & 0.1\%        & 56.0\%        & \textbf{66.7\%} & 62.0\%          
\end{tblr}}
\end{table}

The deep hashing models are also used for multi-category retrieval systems. Therefore, we also evaluated the effectiveness of our attack method \textit{PADHASH} on multi-category datasets. We select the MS-COCO~\cite{lin2015microsoft} as the training dataset for the deep hashing model. The MS-COCO is a multi-category dataset that includes 80 categories, and we randomly select 100,000 images as the training dataset to train the target model.  We randomly select 20\% data of the training dataset to train the surrogate model. As shown in Table \ref{table2-1}, within the poison ratio of 0.1\%, the ASR of our attack method is all above 60\%, which demonstrates that our attack method is still effective for attacking multi-category deep hash models. Therefore, our attack method \textit{PADHASH} is applicable to multi-category retrieval systems in the real world. In addition, compared to the baseline ``Witches' brew'', using our method to generate poisoned can improve the ASR, which verifies the effectiveness of \emph{Strict Gradient-Matching} in the multi-category dataset.

\begin{table*}[t]
\centering
\caption{The performance of data poisoning attack against deep hashing in black-box scenario.}
\begin{tblr}{
  row{odd} = {c},
  row{4} = {c},
  cell{1}{1} = {r=2}{},
  cell{1}{2} = {r=2}{},
  cell{1}{3} = {c=6}{},
  hline{1,5} = {-}{0.08em},
  hline{2} = {3-8}{},
  hline{3} = {-}{},
}
Surrogate Model & Hash Method & Victim Deep Hashing Model &        &          &          &              &        \\
                &             & \textbf{ResNet34}                  & \textbf{VGG11}  & ResNet18 & ResNet50 & MobileNet-v2 & VGG16  \\
Ensemble Model  & CSQ         & 36.0\%                    & 20.0\% & 74.6\%   & 68.6\%   & 88.0\%       & 22.0\% \\
Ensemble Model  & DPN         & 35.3\%                    & 16.0\% & 75.3\%   & 62.0\%   & 85.3\%       & 22.0\% 
\end{tblr}

\label{tab:black}
\end{table*}

In addition, we also conduct experiments on ImageNet1000 mini, whose training dataset contains 100,000 images in 1000 categories. We evaluate the ASR of \textit{PADHASH} on CSQ and DPN on this dataset. The $\alpha$ is 0.2 for both CSQ and DPN and the surrogate dataset is 20\% of the training dataset of ImageNet1000.
As shown in Table \ref{table1-4}, the ASR of our attack method against CSQ and DPN on ImageNet1000 is all above 60\%. For CSQ, the ASR of our attack even exceeds 90\%, which demonstrates the effectiveness of our attack method on large-scale datasets with multi-labels.

\begin{table}[h]
\centering
\caption{The attack success rate of our attack method ImageNet1000. }
\label{table1-4}
\begin{tblr}{
  cells = {c},
  cell{1}{1} = {r=2}{},
  cell{1}{2} = {r=2}{},
  cell{1}{3} = {r=2}{},
  cell{1}{4} = {c=2}{},
  hline{1,4} = {-}{0.08em},
  hline{2} = {4-5}{},
  hline{3} = {-}{},
}
Dataset      & {Hash\\Codes} & Poison Ratio & Attack Success Rate &        \\
             &               &              & CSQ                 & DPN    \\
ImageNet1000 & 128bits       & 0.05\%       & 93.6\%              & 62.4\% 
\end{tblr}
\end{table}

\subsection{Feasibility}
As shown in Table \ref{tab:black}, we evaluate the feasibility of \textit{PADHASH} in the black-box scenario, where the attacker is unaware of the victim model's structure. Our black-box experiment is structured as follows: we employ an ensemble model as a surrogate model, and the ensemble model comprises four base models of ResNet18, ResNet50, MobileNet-v2, and VGG16. The victim deep hashing models are detailed in Table \ref{tab:black}.
When the base model of the victim deep hashing model is ResNet34 or VGG11, we observe ASR is approximately 35\% and 20\%, respectively, indicating that attackers can still potentially mount successful attacks in a practical scenario and demonstrating the feasibility of \textit{PADHASH}. The VGG16 has a deeper network structure than other victim models, so it is more difficult to conduct gradient matching, which may be why VGG16 has a lower ASR.

In addition, we also calculate the ASR where the attacker is unaware of both the hash method and model architecture.  In Table \ref{transfer2}, the surrogate model is an ensemble model that is the same as above. We can observe that the average ASR is 30\% even though the surrogate model and surrogate hash method are different from the target model and target hash method, which demonstrates the feasibility of \textit{PADHASH}.

\begin{table}[h]
\centering
\caption{The ASR of \textit{PADHASH} where the attacker is unaware of both the hash method and model architecture.}
\begin{tblr}{
  row{odd} = {c},
  row{4} = {c},
  cell{1}{1} = {r=2}{},
  cell{1}{2} = {r=2}{},
  cell{1}{3} = {r=2}{},
  cell{1}{4} = {c=2}{},
  hline{1,5} = {-}{0.08em},
  hline{2} = {4-5}{},
  hline{3} = {-}{},
}
Surrogate Model & {Surrogate \\Method } & {Target \\Method } & Victim Model &        \\
                &                       &                    & ResNet34     & VGG 11 \\
Ensemble Model  & CSQ                   & DPN                & 48.0\%       & 20.6\% \\
Ensemble Model  & DPN                   & CSQ                & 35.3\%       & 16.6\% 
\end{tblr}
\label{transfer2}
\end{table}

\subsection{Comparative Analysis}
Although \textit{PADHASH }is the first data poisoning attack against deep hashing models, there are other attack methods, such as backdoor attacks and clean image attacks, that are similar to ours. In this section, we will compare the effectiveness of \textit{PADHASH} and these related attacks. 
Specifically, the similar attack method based on clean images we compared is Clean-Image Backdoor~\cite{chen2022clean}, and the related attack method against deep hashing models we compared is BadHash~\cite{hu2022badhash}. In addition, we also select BadNet~\cite{gu2017badnets} as the baseline for comparison.

\textit{1) Compared With Backdoor-based Attacks.}  We first evaluate the ASR of our attack method and backdoor-based attacks on the ImageNet100 and MS-COCO. The hash codes are 64bit, and the base model is ResNet50. The following is the specific experimental setting of these attacks.
\begin{itemize}
    \item BadHash: We follow the experimental setting of BadHash\cite{hu2022badhash}. The normalized trigger is limited to (-8/255, 8/255). The poisoned images were randomly selected from the training dataset, and their labels were changed to the target label 42. 
    \item BadNet: We use randomly generated noise normalized to (0,1) as the trigger pattern. The length and width of the trigger are both 8, and the trigger is attached to the lower right corner of the poisoned image. The examples of poisoned images are shown in Figure \ref{fig:steal1}, and their labels were also changed to the target label 42
\end{itemize}

As shown in Table \ref{table1-1}, we evaluate the ASR using different poison ratios: 0.05\%, 0.1\%, and 0.5\%. For the ImageNet100, our attack method has higher ASR than BadHash and BadNet for both CSQ (78.2\% vs. 1.1\% vs. 0.8\%) and DPN (82.6\% vs. 20.2\% vs. 0.5\%) at a poisoning rate of 0.05\%.  Similarly, for the MS-COCO, our attack method also has higher ASR for both CSQ (82.6\% vs. 2.0\% vs. 0.8\%) and DPN (62.0\% vs. 1.3\% vs. 0.5\%) at a poisoning rate of 0.1\%. The baseline is effective when the poison ratio is twice or even more than that of our attack method.
This finding demonstrates that our attack method is more effective than baselines under a lower poison ratio, which is beneficial for attacking the deep hashing models since it is easier for attackers to poison the training dataset.

\begin{table}[h]
\centering
\caption{The ASR of our attack method and baselines under different poison ratios. The hash codes are 64bit, and the base model is ResNet50.}
\label{table1-1}
\scalebox{0.78}{
\begin{tblr}{
  cells = {c},
  cell{1}{1} = {r=2}{},
  cell{1}{2} = {r=2}{},
  cell{1}{3} = {c=3}{},
  cell{1}{6} = {c=3}{},
  cell{3}{1} = {r=3}{},
  cell{6}{1} = {r=2}{},
  hline{1,8} = {-}{0.08em},
  hline{2} = {3-8}{},
  hline{3,6} = {-}{},
}
Dataset     & {Poison \\Ratios} & CSQ    &         &                  & DPN    &         &                          \\
            &                   & BadNet & BadHash & Ours             & BadNet & BadHash & Ours                     \\
ImageNet100 & 0.05\%            & 0.8\%  & 1.1\%   & \textbf{78.2\%}  & 0.5\%  & 20.2\%  & \textbf{\textbf{82.6\%}} \\
            & 0.1\%             & 1.1\%  & 1.3\%   & \textbackslash{} & 1.6\%  & 63.7\%  & \textbackslash{}         \\
            & 0.5\%             & 88.0\% & 82.6\%  & \textbackslash{} & 97.6\% & 95.2\%  & \textbackslash{}         \\
MS-COCO     & 0.1\%             & 0.26\% & 2.0\%   & \textbf{82.6\%}  & 6.6\%  & 1.3\%   & \textbf{\textbf{66.7\%}} \\
            & 0.2\%             & 32.0\% & 86.0\%  & \textbackslash{} & 24.8\% & 84.0\%  & \textbackslash{}         
\end{tblr}}
\end{table}

\begin{figure}[h]
\centering
    \centering
    \includegraphics[width=0.45\textwidth]{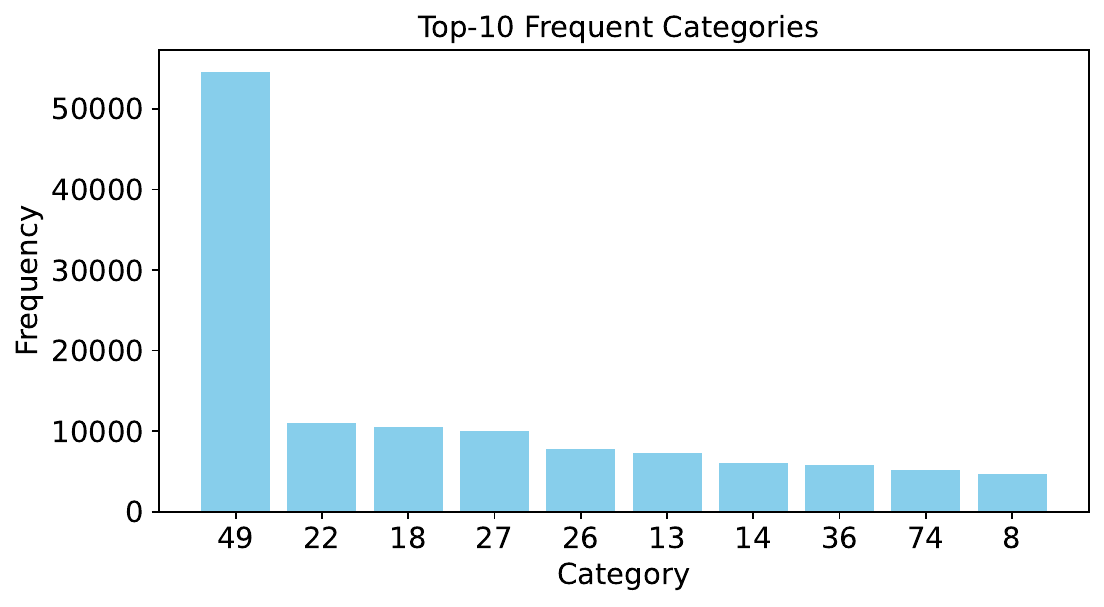}
    \caption{The top-10 frequency categories in MS-COCO. }
    \label{fig:4}
\end{figure}

\textit{2) Compared with \textbf{C}lean \textbf{I}mage \textbf{B}ackdoor Attack (CIB).}
We evaluate its effectiveness on the MS-COCO dataset. We first count the frequency of different categories in the dataset, and the top-10 frequency categories are detailed in Figure \ref{fig:4}. We found that among images with three or more categories, the number of images labeled [49, 22, 27,...] is the largest, where [49, 22, 27,...] represents the image contains at least three labels: 49, 22, 27. Therefore, we select images labeled [49, 22, 27,...] and add a new label, 42, to obtain poisoned images labeled [49, 22, 27, 42, ...]. We define this attack as the Clean Image Backdoor Add attack (\textbf{CIB-Add}). In addition, we also performed a Clean Image Backdoor Replace attack (\textbf{CIB-Rep}). Specifically,  if an image is labeled with [1,...] and does not contain label 0, we alter its label to [0,...]. 

As shown in Table~\ref{table1-2}, we evaluate the above two attacks, CIB-Add and CIB-Rep, and our attack method on the MS-COCO dataset. The results show that no matter whether the threshold is 10\% or 30\%, the ASR of our attack method is much higher than CIB-Add and CIB-Rep at a poisoning rate of 0.1\%, which indicates that our method is more suitable for attacking deep hash models than clean image backdoor attacks.

\begin{table}[h]
\centering
\caption{The ASR of our attack method and Clean Image Backdoor under different poison ratios and metric thresholds on the MS-COCO dataset.}
\label{table1-2}
\scalebox{0.8}{
\begin{tblr}{
  cells = {c},
  cell{1}{1} = {r=2}{},
  cell{1}{2} = {r=2}{},
  cell{1}{3} = {c=3}{},
  cell{1}{6} = {c=3}{},
  cell{3}{1} = {r=4}{},
  cell{7}{1} = {r=4}{},
  hline{1,11} = {-}{0.08em},
  hline{2} = {3-8}{},
  hline{3,7} = {-}{},
}
{Hash\\Method} & {Poison\\Ratio} & Threshold: 30\% &         &                  & Threshold: 10\% &         &                  \\
            &              & CIB-Add         & CIB-Rep & Ours             & CIB-Add         & CIB-Rep & Ours             \\
CSQ         & 0.0\%        & 0.6\%           & 0.2\%   & \textbackslash{} & 16.8\%          & 0.2\%   & \textbackslash{} \\
            & 0.1\%        & 0.6\%           & 5.8\%   & \textbf{79.3\%}  & 16.8\%          & 6.2\%   & \textbf{86.0\%}  \\
            & 1\%          & 0.6\%           & 44.4\%  & \textbackslash{} & 16.8\%          & 47.1\%  & \textbackslash{} \\
            & 2\%          & 0.6\%           & 0.6\%   & \textbackslash{} & 16.8\%          & 76.2\%  & \textbackslash{} \\
DPN         & 0.0\%        & 1.1\%           & 0.1\%   & \textbackslash{} & 24.4\%          & 0.1\%   & \textbackslash{} \\
            & 0.1\%        & 1.1\%           & 3.6\%   & \textbf{66.7\%}  & 24.4\%          & 4.2\%   & \textbf{73.3\%}  \\
            & 1\%          & 1.1\%           & 41.2\%  & \textbackslash{} & 24.4\%          & 44.9\%  & \textbackslash{} \\
            & 2\%          & 1.1\%           & 0.8\%   & \textbackslash{} & 24.4\%          & 73.2\%  & \textbackslash{} 
\end{tblr}}
\end{table}

In addition, we find some interesting results for both CIB-Add and CIB-Rep.
For the CIB-Add attack, the ASR under all poisoning rates is almost the same.  We infer that adding a new label to the poisoned image has little impact on the training of the deep hashing model. For instance, if an image is labeled [people, dog,..., car] and we add a new label ``cat'' to obtain the poisoned image labeled [people, dog, \textbf{cat}, ..., car], the model may prefer to learn the features of [people, dog,..., car] rather than learning the features of ``cat'' since there are no features of ``cat'' in the poisoned image for the model to learn. For the CIB-Rep attack, when the threshold is 30\%, the ASR of 1\% poison ratio is higher than that of 2\%. This is because when the poisoning rate is 2\%, the Category 0 and Category 1 images are close in the Hamming space. In the MS-COCO dataset, there are 2494 images with label 0 and 1344 images with label 1. Therefore, when querying an image of category 0, many images of category 0 will be retrieved, resulting in a lower ASR when the threshold is 30\%.

\textit{3) Qualitative Comparison}. We have quantitatively compared the effectiveness of our method with related methods in the above content. In this section, we will qualitatively compare our method with related methods. As shown in Table \ref{table1-3}, we provide a summary table to compare our method with related methods. The following briefly explains the Table \ref{table1-3}. 

\begin{table}[h]
\centering
\caption{Comparison of our attack method and related attack methods. }
\label{table1-3}
\begin{tblr}{
  cells = {c},
  hline{1,5} = {-}{0.08em},
  hline{2} = {-}{},
}
Method  & Clean-Label  & Clean Trigger Image & Poisoned Image         \\
CIB     & No           & \textbf{Yes}        & \textbf{Clean Image} \\
BadHash & No           & No                  & Perturbed Image      \\
Ours    & \textbf{Yes} & \textbf{Yes}       & Perturbed Image      
\end{tblr}
\end{table}

\begin{itemize}
    \item Clean-Label: For the Clean-Image Backdoor and BadHash, they need to modify the label of the poison image while Our method does not need.
    \item Clean Trigger Image: In the attack stage, BadHash needs to add a trigger to the clean image to attack the deep hashing model, while our method and Clean-Image Backdoor can use clean images to attack.
    \item Poisoned Image: BadHash and our method need to perturb the image to obtain the poison image, while the poison image of Clean-Image Backdoor is the clean image.
\end{itemize}

Clean-label and clean trigger image make our attack method stealthy when poisoning and attacking the deep hashing model. 
Although our method needs to perturb the clean image to obtain poisoned images, the perturbation added to the clean image can be limited to 8/255 after normalization. In addition, we can increase the limit on perturbation for images with larger sizes to make them more invisible. 
Based on the above analysis and the experimental results, our attack method is generally more stealthy and performs better during the attack.

\begin{table*}[t]
\centering
\caption{The result of data poisoning attack on integrity. MAP and MAP* are the mean average precise of the clean and deep hashing model.}
\begin{tblr}{
  cells = {c},
  cell{2}{1} = {r=2}{},
  cell{4}{1} = {r=2}{},
  hline{1,6} = {-}{0.08em},
  hline{2} = {-}{},
}
Hash Method & Dataset     & Poison Num & Poison Ratio & Test Image Num & Hash Codes & MAP$\uparrow$    & MAP*$\uparrow $   \\
CSQ         & CIFAR10     & 100        & 0.25\%       & 2000           & 32bits     & 83.1\% & 83.7\% \\
            & ImageNet100 & 65         & 0.05\%       & 1000           & 64bits     & 78.6\% & 79.1\% \\
DPN         & CIFAR10     & 100        & 0.25\%       & 2000           & 32bits     & 83.4\% & 83.5\% \\
            & ImageNet100 & 65         & 0.05\%       & 1000           & 64bits     & 77.8\% & 77.5\% 
\end{tblr}
\label{inter}
\end{table*}

\subsection{Integrity}
Preserving the model's integrity is crucial since it makes it challenging for a model trainer to discern whether a deep hashing model has been compromised, increasing the likelihood of the compromised model being deployed. Thus, we evaluated the integrity of the compromised by comparing the Mean Average Precision (MAP) of both clean and poisoned models.
As indicated in Table \ref{inter}, the MAP values of the compromised models are similar to those of the clean models and did not decrease significantly. The results indicate that \textit{PADHASH} preserves the integrity of the deep hashing model, thereby facilitating the covert execution of the attack. 

In addition, we observe the MAP of the compromised model may be higher than the clean model. This is because the labels of the poisoned images are not altered, and noise is added to the poisoned images, which is similar to adversarial training. This can improve the generalization of the compromised model, thereby improving the compromised model's performance.

\subsection{Stealthiness} In the process of injecting the poisoned images, it is imperative to ensure concealment of them. Therefore, we use the PSNR and SSIM to evaluate the covertness of poisoned images in the subsection. As detailed in Table \ref{tab:stealthiness}, the poisoned images exhibit SSIM value above 0.9 and PSNR close to 30, indicating they remain visually similar to the original images. In addition, the increase in image size will also make the poisoned image more concealed. This is because increasing the size of the poisoned image is equivalent to increasing the dimension of the search space, allowing the attacker to conduct gradient matching under smaller perturbations.

\begin{table}[h]
\caption{PSNR and SSIM between poisoned and clean images.}
\centering
\begin{tblr}{
  cells = {c},
  hline{1,6} = {-}{0.08em},
  hline{2} = {-}{},
}
Hase Method & Dataset  & PSNR$\uparrow $  & SSIM$\uparrow $  \\
CSQ         & CIFAR10  & 30.35 & 0.909 \\
CSQ         & ImageNet100 & 36.79 & 0.944 \\
DPN         & CIFAR10  & 29.45 & 0.901 \\
DPN         & ImageNet100 & 36.75 & 0.919      
\end{tblr}

\label{tab:stealthiness}
\end{table}

\subsection{Transferability}

\begin{figure}[h]
\centering
    \centering
    \includegraphics[width=0.3\textwidth]{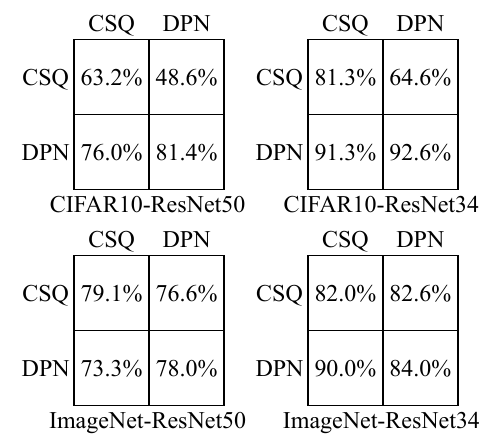}
    \caption{The ASR of \textit{PADHASH} across deep hashing methods.}
    \label{fig:tran}
\end{figure}

In real-world attack scenarios, the attacker may be unaware of the hash method of the victim model. Therefore, we conducted transferability experiments on different hash methods in this subsection to explore whether \textit{PADHASH} has transferability between different hash methods.
As depicted in Figure \ref{fig:tran}, the horizontal axis denotes the victim hash method, and the vertical axis represents the surrogate hash method the attacker uses to train the surrogate model and generate the poisoned images. The results show that the ASR exceeds 60\% in most transfer attack settings, demonstrating the transferability of \textit{PADHASH} across different deep hashing methods and highlighting its potential practical effectiveness.

\subsection{Attack Robustness.}
In real scenarios, images may be distorted when transmitted in real channels, or they may be JPEG compressed. Therefore, we need to explore whether \emph{PADHASH} is robust to some common distortions. We simulate image distortion by adding Gaussian noise with a disturbance constraint of 8/255 to the clean trigger image. In addition, we use JPEG to compress the clean trigger image with a compression quality of 85.
\begin{table}[h]
\centering
\caption{ASR of \emph{PADHASH} under Gaussian noise and JPEG compression attack.}
\label{robust}
\begin{tblr}{
  row{even} = {c},
  cell{1}{1} = {r=2}{},
  cell{1}{2} = {r=2}{c},
  cell{1}{3} = {c=3}{c},
  cell{3}{1} = {r=2}{},
  cell{3}{2} = {c},
  cell{3}{3} = {c},
  cell{3}{4} = {c},
  cell{3}{5} = {c},
  hline{1,5} = {-}{0.08em},
  hline{2} = {3-5}{},
  hline{3} = {-}{},
}
Dataset & Hash Method & Attack Success Rate~ &                &        \\
        &             & Original             & Gaussian noise & JPEG   \\
CIFAR10 & CSQ         & 69.0\%               & 70.0\%         & 65.0\% \\
        & DPN         & 76.0\%               & 81.0\%         & 77.0\% 
\end{tblr}
\end{table}
As shown in Table \ref{robust}, when the clean trigger image is added with Gaussian noise or JPEG compressed, the attack success rate is still close to the original image, which demonstrates that \emph{PADHASH} is robust to image distortion and JPEG compression in real scenarios. Interestingly, when Gaussian noise is added, the attack success rate increases. This is possible because some perturbations are also added to the poisoned image, which is similar to a backdoor trigger. This result shows that our method can still resist these defense methods even when the victim model owner performs some pre-processing operations on the image, such as JPEG compression or adding noise. 
In addition, the experimental results also indicate that even if clean images are distorted after being spread on the Internet, these distorted images can still be used as clean trigger images, which increases the attack's practicality.

\begin{table*}[t]
\centering
\caption{The ASR across different values of hyperparameter $\alpha$.}
\label{alpha}
\begin{tblr}{
  cells = {c},
  cell{1}{1} = {r=2}{},
  cell{1}{2} = {r=2}{},
  cell{1}{3} = {c=9}{},
  cell{3}{1} = {r=2}{},
  cell{5}{1} = {r=2}{},
  hline{1,7} = {-}{0.08em},
  hline{2} = {3-11}{},
  hline{3} = {-}{},
}
Dataset     & Hash Method & Hyperparameter~$\alpha$ &        &        &        &        &        &        &        &        \\
            &             & 0.0                     & 0.05   & 0.1    & 0.15   & 0.2    & 0.25   & 0.3    & 0.35   & 0.4    \\
CIFAR10     & CSQ         & 44.4\%                  & 44.4\% & 50.4\% & 56.0\% & 67.6\% & 59.2\% & 61.6\% & 66.8\% & 64.0\% \\
            & DPN         & 80.0\%                  & 81.4\% & 74.4\% & 76.4\% & 74.0\% & 70.8\% & 69.6\% & 66.0\% & 57.2\% \\
ImageNet100 & CSQ         & 24.0\%                  & 37.3\% & 46.0\% & 56.0\% & 64.0\% & 66.6\% & 72.6\% & 74.0\% & 76.0\% \\
            & DPN         & 78.0\%                  & 81.3\% & 85.3\% & 89.3\% & 87.3\% & 97.5\% & 90.0\% & 88.0\% & 88.0\% 
\end{tblr}
\end{table*}

\subsection{Ablation Study}

\textit{1) Hyperparameters.}
In section \ref{metho}, we introduce \textit{PADHASH} for attacking deep hashing models. There is an important parameter $\alpha$ in this method for balancing two losses, and choosing a suitable $\alpha$ is critical for \textit{PADHASH}. 
As shown in Table \ref{alpha}, we calculate the ASR across different hyperparameter values $\alpha$. We can observe that increasing $\alpha$ within a certain range will increase ASR, but if $\alpha$ exceeds a certain threshold, ASR will decrease.

Notably, the choice of $\alpha$ is related to the loss function of the target deep hashing model. For a smoother loss function, $\alpha$ can be larger. Otherwise, $\alpha$ should be smaller. For instance, compared with CSQ, DPN needs to choose a smaller $\alpha$ since the CSQ uses binary cross-entropy loss, while DPN uses polarization loss, which is steeper than cross-entropy loss. In addition, for images of larger size, a larger alpha can be chosen. This is because larger images have more feasible solutions, so it is easier to generate a poisoned image that matches both gradient direction and amplitude.

\textit{2) Surrogate Dataset.}
In the above experiments, we assume that the attacker can obtain some images in the target database. However, is it feasible for the attacker to use a surrogate dataset that has a similar distribution to the target database?

Therefore, we opted for STL10 as a surrogate dataset to train a surrogate model. Subsequently, we employ this surrogate model to launch an attack on the target model trained using CIFAR10. As shown in Table \ref{sur}, the ASR of all attack settings exceeds 25\%. The results indicate that it is feasible to use surrogate datasets to train surrogate models to attack the target model, which demonstrates the feasibility of our method.

\begin{table}[h]
\centering
\caption{The ASR of our method when using surrogate datasets to attack.}
\begin{tblr}{
  cells = {c},
  hline{1,4} = {-}{0.08em},
  hline{2} = {-}{},
}
Method & Poison Ratio & Baseline-ASR & Ours-ASR        \\
CSQ    & 0.25\%       & 41.9\%($\pm$ 3.69)       & \textbf{46.9\%}($\pm$ 5.66) \\
DPN    & 0.25\%       & 26.8\%($\pm$ 1.85)       & \textbf{28.7\%}($\pm$ 2.01)
\end{tblr}
\label{sur}
\end{table}

\textit{3) Base Model.}
In our study, the deep hashing model is constructed on top of a Deep Neural Network (DNN) model, referred to as the base model. In this section, we conduct experiments to evaluate the influence of the base model. 
The results in Table \ref{base model} indicate that our \textit{PADHASH} method maintains an attack success rate exceeding 55\% in all base models, demonstrating the generality of \textit{PADHASH} across different base models. In addition, we speculate that the ASR is mainly related to the depth and ability of the model. For the same type of model, increasing its depth will increase the difficulty of gradient matching, but it will also strengthen the feature extraction ability of the model. This may be why the ASR of ResNet34 is higher than that of ResNet50 and ResNet18.

\begin{table}[h]
\centering
\caption{The impact of the base model on ASR.}
\begin{tblr}{
  cells = {c},
  cell{2}{1} = {r=4}{},
  cell{6}{1} = {r=4}{},
  hline{1,10} = {-}{1pt},
}
Hash Method & Dataset       & Model & Poison Ratio & ASR$\uparrow $    \\
\hline
CSQ         & CIFAR10       & ResNet34 & 0.25\%    & 81.3\% \\
            & ImageNet100   & ResNet34 & 0.05\%    & 82.0\% \\
            & CIFAR10       & ResNet18 & 0.25\%    & 74.0\% \\
            & ImageNet100   & AlexNet  & 0.05\%    & 55.3\% \\
\hline
DPN         & CIFAR10       & ResNet34 & 0.25\%    & 92.6\% \\
            & ImageNet100   & ResNet34 & 0.05\%    & 84.0\% \\
            & CIFAR10       & ResNet18 & 0.25\%    & 86.0\% \\
            & ImageNet100   & AlexNet  & 0.05\%    & 74.0\% 
\end{tblr}
\label{base model}
\end{table}

\begin{figure}[h]
\centering
    \centering
    \includegraphics[width=0.4\textwidth]{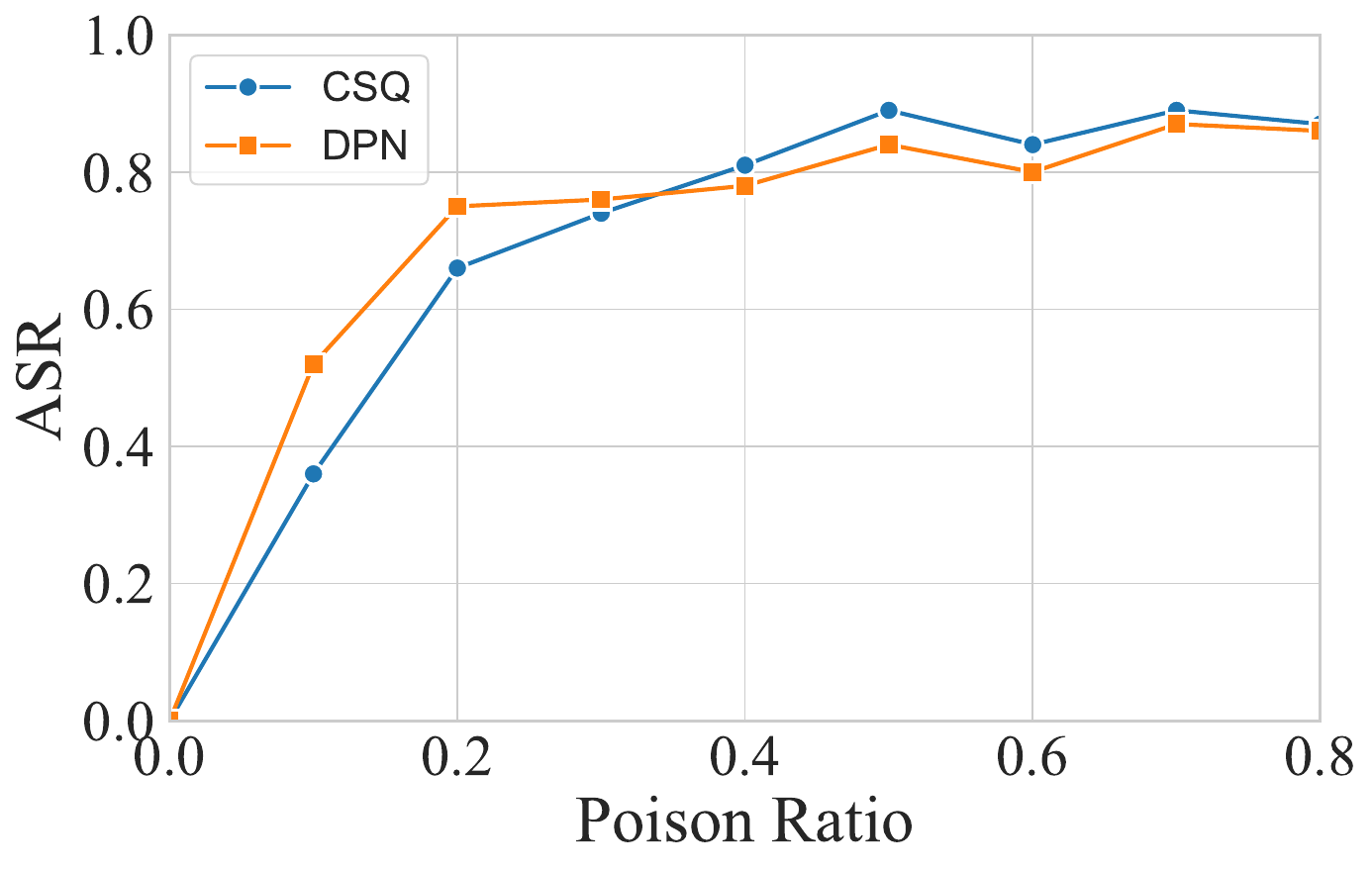}
    \caption{The impact of poison ratio on ASR. The dataset is CIFAR10 and the base model is ResNet50.}
    \label{fig:rate}
\end{figure}

\textit{4) Poison Ratio.}
As shown in Figure \ref{fig:rate}, we study the impact of the poison ratio on the ASR. When the poisoning ratio is less than 0.2\%, the poisoning ratio greatly impacts ASR, and increasing the number of poisoned images improves the ASR. When the poisoning ratio reaches 0.2\%, the ASR can exceed 60\%, validating that \textit{PADHASH} is effective at a low poisoning ratio. When the poisoning rate is higher than 0.2\%, the ASR tends to be stable. This is because ASR is mainly affected by other factors such as gradient matching, base model, perturbation constraints, etc.

\textit{5) Hash Codes Length. }To investigate the impact of hash codes on the success rate of attacks, we study the attack success rate under different hash code lengths.  
As shown in Table \ref{table6}, although the hash code length of the surrogate model is different from the victim model, the ASR can exceed 60\% in all settings, which demonstrates the transferability of \textit{PADHASH} across models with different hash codes lengths, making \textit{PADHASH} more practical in a real-world attack. In addition, the reason why \textit{PADHASH} has transferability in hash codes is that the clean trigger images and the malicious images are similar in the feature space, even though the hash code lengths are different.

\begin{table}[h]
\centering
\caption{The impact of hash codes on ASR. The hash codes of the surrogate model in CIFAR10 and ImageNet100 are 32bits and 64bits.}
\begin{tblr}{
  cells = {c},
  cell{1}{1} = {r=2}{},
  cell{1}{2} = {c=3}{},
  cell{1}{5} = {c=3}{},
  hline{1,5} = {-}{0.08em},
  hline{2} = {2-7}{},
  hline{3} = {-}{},
}
{Hash \\Method } & CIFAR10 &        &        & ImageNet100 &        &         \\
                 & 16bits  & 32bits & 64bits & 32bits   & 64bits & 128bits \\
CSQ              & 60.4\%  & 67.6\% & 70.0\% & 79.3\%   & 79.1\% & 69.3\%  \\
DPN              & 68.7\%  & 76.0\% & 65.4\% & 85.3\%   & 78.0\% & 82.6\%  
\end{tblr}
\label{table6}
\end{table}

\textit{5) Metric Threshold.}
We set a threshold to measure whether an attack is successful. In the above experiment, we set the threshold to 0.3, which means that the attack is considered successful only when the target category images account for more than 30\% of the retrieved images. To eliminate the impact of threshold selection on the experimental results, we counted the attack success rates under different thresholds.

As shown in Table \ref{threshold}, as the threshold increases, the attack success rate of our method and the baseline decreases. However, our method has a higher ASR under various threshold selections, which demonstrates that the threshold selection will not affect the conclusion that our method can improve the ASR.

\begin{table*}[t]
\centering
\caption{The ASR under different threshold.}
\begin{tblr}{
  row{odd} = {c},
  row{4} = {c},
  row{6} = {c},
  cell{1}{1} = {r=2}{},
  cell{1}{2} = {r=2}{},
  cell{1}{3} = {c=9}{},
  cell{3}{1} = {r=2}{},
  cell{5}{1} = {r=2}{},
  hline{1,7} = {-}{0.08em},
  hline{2} = {3-11}{},
  hline{3} = {-}{},
  hline{5} = {2-11}{},
}
Hash Method & Method   & Metric Threshold &        &        &        &        &        &        &        &        \\
            &          & 10\%             & 20\%   & 30\%   & 40\%   & 50\%   & 60\%   & 70\%   & 80\%   & 90\%   \\
CSQ         & Baseline & 52.0\%           & 47.6\% & 44.4\% & 43.6\% & 42.8\% & 42.4\% & 41.6\% & 38.8\% & 34.8\% \\
            & Ours     & 72.0\%           & 69.2\% & 67.6\% & 65.6\% & 65.6\% & 64.0\% & 62.8\% & 60.4\% & 53.2\% \\
DPN         & Baseline & 82.4\%           & 81.4\% & 80.0\% & 79.4\% & 78.6\% & 78.2\% & 78.0\% & 77.4\% & 74.2\% \\
            & Ours     & 82.8\%           & 82.0\% & 81.4\% & 80.2\% & 79.6\% & 79.6\% & 78.6\% & 77.8\% & 74.2\% 
\end{tblr}
\label{threshold}
\end{table*}

\begin{figure*}[t]
\centering
    \centering
    \includegraphics[width=0.9\textwidth]{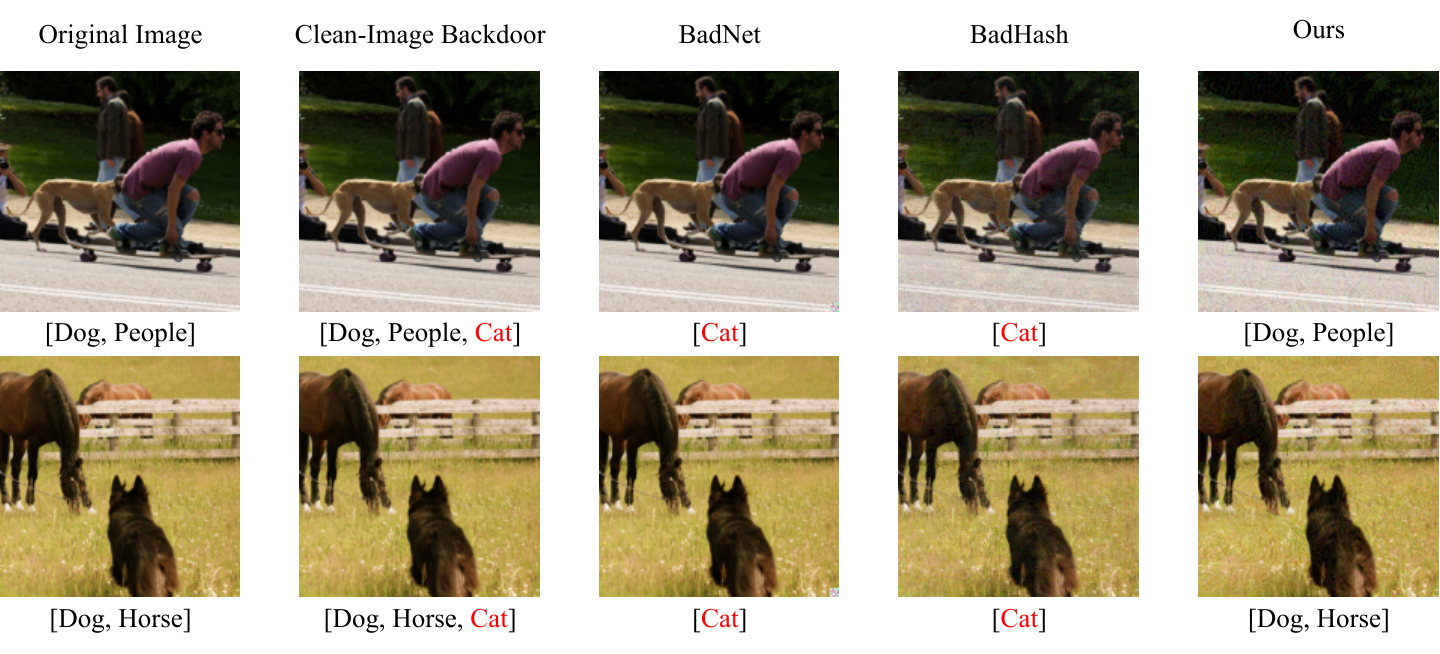}
    \caption{Visualization results of original and poisoned images of our attack method and related attack methods.}
    \label{fig:steal1}
\end{figure*}

\subsection{Visualization}
Concealment translates to heightened difficulty in detecting these poisoned data instances. Therefore, we present a selection of sample poisoned images of our attack method and related attack methods, as depicted in Figure \ref{fig:steal1}. Remarkably, these visual representations reveal an exceedingly subtle distinction between the poisoned and original images sourced from the MS-COCO dataset, which demonstrates that the poisoned images generated by \emph{PADHASH} are also visually concealed. In addition, we emphasize that in real-world attack scenarios, the poisoned images we generate will be more visually invisible. Because the images in the real world are larger, and the perturbation search space is larger. Therefore, gradient matching can be effectively accomplished within more constrained perturbation bounds, contributing to the minimal visual divergence observed. Moreover, the poisoned images of \emph{PADHASH} consist of their semantic labels, while those of Clean Image Backdoor, BadNet, and BadHash do not. 

\section{Discussion}

\subsection{Defensive Strategy}

\emph{PADHASH} demonstrate that clean images may also be dangerous, which reveals the potential risks of deep hashing models. In practical applications, the purpose of an attack is to allow the defender to find the weaknesses of the system and implement corresponding defenses. Therefore, we need to discuss potential defense strategies against PADHASH, mainly divided into poisoned data detection, data augmentation, and robust training.

\textit{1) Poisoned Data Detection.} Although our poisoned images are clean-label, they may differ from clean images in feature space. Therefore, the defender may be able to detect poisoned images in feature space, such as clustering~\cite{detect}. Once the poisoned images are detected and removed, the security of the deep hash model can be effectively protected.

\textit{2) Data Augmentation.} Data augmentation is a general defense method for data poison attacks. For instance, we can add Gaussian noise and crop the images in the training set to make the poisoned images invalid. However, when performing data augmentation operations, it is necessary to consider the trade-off between the quality of the augmented dataset and defensive performance.

\textit{3) Robust training.} Robust training~\cite{jia2021intrinsic} divides the training set into multiple subsets and then uses the multiple subsets to train multiple models to obtain the ensemble model. Since each subset is randomly sampled, the number of poisoned images in each subset will be reduced, alleviating the effectiveness of data poisoning attacks. However, if the dataset is divided into too many subsets, the performance of the model trained on each subset may be poor, which will also result in poor performance of the ensemble model. Therefore, it is worth exploring how to achieve a trade-off between defense performance and model performance.

\subsection{Potential Application.}
Although \textit{PADHASH} is an attack on multimedia retrieval systems, it still has broader implications for multimedia systems. First, our attack method can reveal the potential risks of deep hashing models. This is equivalent to \textit{penetration testing} to explore the weaknesses of the current multimedia retrieval system. Then we can design targeted defense methods to defend against attacks, which is also one of the goals of our research on this attack method. 

In addition, \textit{PADHASH} can also benefit multimedia retrieval systems. For instance, \textit{PADHASH} can be transferred to a model fingerprint to protect the copyright of multimedia retrieval systems. Specifically, we first select some clean trigger images and select a target category. Then, we generate poisoned images for these clean trigger images that match the gradient of images of the target category. Finally, we can use poisoned images to attack our deep hashing model to embed the fingerprint. When we need to verify the copyright of a suspected model, we input clean trigger images to the suspected model. If we can retrieve images of the target category, we can confirm that the suspected model is ours.

\subsection{Computational Consumption}
In practical applications, the computational consumption of an attack has an important impact on the application of the attack.
The computational resource consumption of \textit{PADHASH} is mainly for training the surrogate model and generating poisoned images. 

\textit{1) Training Surrogate Model.} The computational consumption of training surrogate model is mainly related to the size of the surrogate dataset. Fortunately, a surrogate dataset that is only 10\% of the target model’s training set is effective enough to train a surrogate model.
Therefore, the time of training the surrogate model is significantly less than the training time of training the attacked model. 
For instance, the training dataset of ImageNet100 includes 130,000 samples. We use an RTX 4090 24GB to train a deep hashing model on this dataset with ten epochs requiring about 20 minutes. However, training a surrogate model only requires about 80 seconds. Even though the training dataset is full ImageNet that includes 1.2$M$ images, we only need 10\% of the training data to train the surrogate model, so the computational consumption is affordable.

\begin{figure}[h]
\centering
    \centering
    \includegraphics[width=0.5\textwidth]{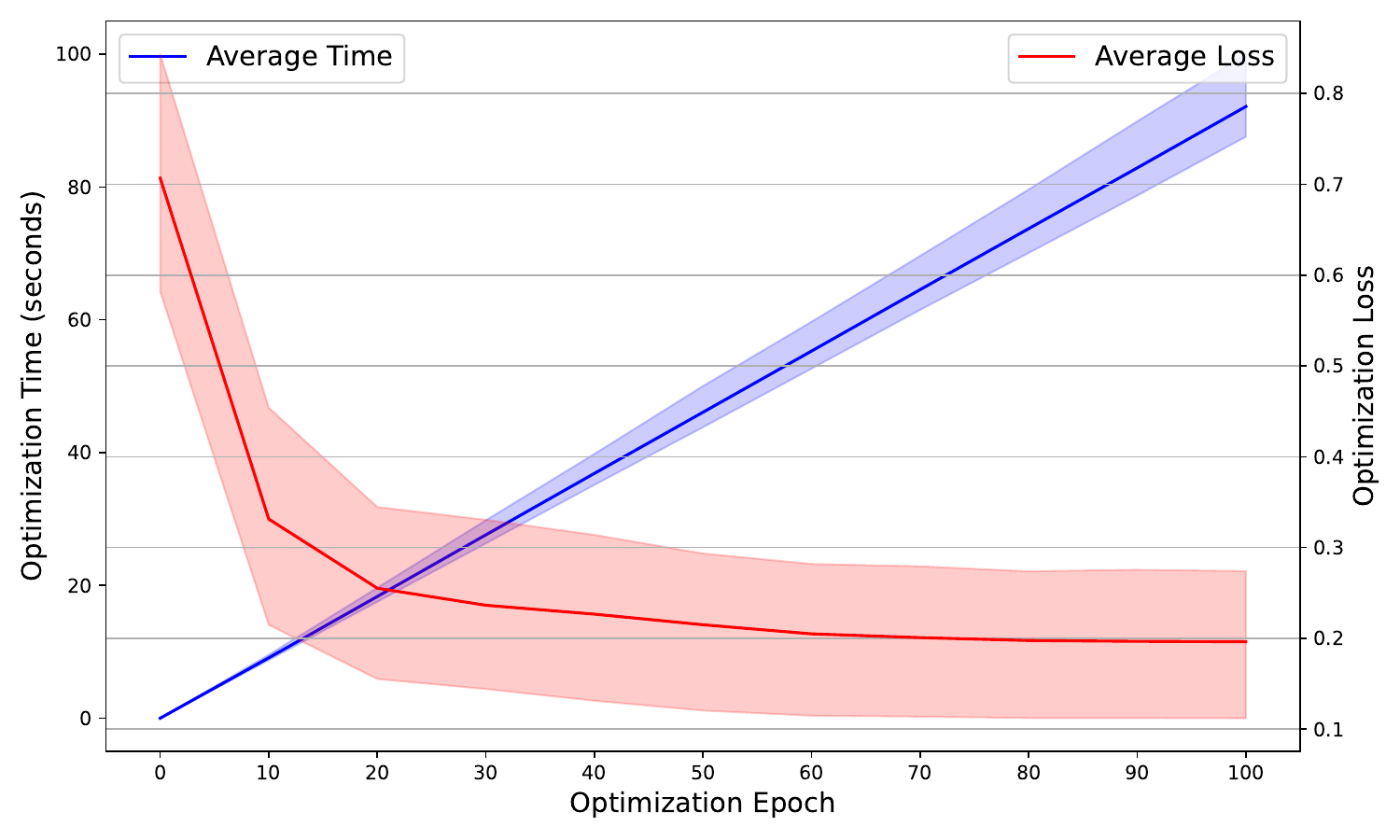}
    \caption{The optimization time and loss per epoch with the Min/Max range. }
    \label{fig:2-1}
\end{figure}

\textit{2) Generate Poisoned Images.} Our attack method requires gradient-matching optimizations when generating poisoned images. We have evaluated the computational consumption of generating the poisoned images. As shown in Figure \ref{fig:2-1}, we optimized 100 poisoned images, and the optimization time increases linearly with the optimization epoch. In our experiments, the poisoned images are optimized for 60 epochs, which requires about 60 seconds. Therefore, the computational consumption of generating poisoned images is also affordable.

\section{Conclusion}
In this paper, we propose the first data poisoning attacks against deep hashing models to explore their potential risks. The attacker generates the poison images to compromise the deep hashing models. When users query with clean trigger images, they will obtain malicious images such as violent, explicit, and private images. Our experiments in gray-box and black-box scenarios validate that our proposed data poisoning attacks against deep hashing models are effective and practical on different datasets and models. In addition, we propose a \emph{Strict Gradient-Matching} method to generate poisoned images, which has been demonstrated to improve the attack success rate. Our proposed attack method reveals the potential risks of the deep hashing model. Therefore, we call on not only paying attention to the performance of deep hashing models but also to the security of deep hashing models.

\section*{Acknowledgments}
This work was supported in part by the National Natural Science Foundation of China under Grant U2336206, 62472398, U2436601 and 62402469. Additionally, this research is supported by the National Research Foundation, Singapore and Infocomm Media Development Authority under its Trust Tech Funding Initiative (No. DTCRGC-04). Any opinions, findings and conclusions or recommendations expressed in this material are those of the author(s) and do not reflect the views of National Research Foundation, Singapore and Infocomm Media Development Authority.

\bibliographystyle{IEEEtran}
\bibliography{TMM}

\end{document}